\documentclass{article}

\PassOptionsToPackage{square,numbers}{natbib}
\usepackage{graphicx}

\usepackage[preprint]{neurips_2023}

\usepackage[utf8]{inputenc} 
\usepackage[T1]{fontenc}    
\usepackage{hyperref}       
\usepackage{url}            
\usepackage{booktabs}       
\usepackage{amsfonts}       
\usepackage{nicefrac}       
\usepackage{microtype}      
\usepackage{xcolor}         
\usepackage{float}
\usepackage{color}
\usepackage{tcolorbox}
\usepackage{pdfpages}
\usepackage{listings}
\tcbuselibrary{listings,breakable}
\title{LLaMandement : \\
Large Language Models for Summarization of French Legislative Proposals.}

\newtcolorbox{mybox}{
  boxrule=0.5pt,
  arc=4pt,
  left=6pt,
  right=6pt,
  boxsep=5pt,
  breakable
}

\definecolor{codegreen}{rgb}{0,0.6,0}
\definecolor{codegray}{rgb}{0.5,0.5,0.5}
\definecolor{codepurple}{rgb}{0.58,0,0.82}
\definecolor{backcolour}{rgb}{0.95,0.95,0.92}

\lstdefinestyle{mystyle}{
    backgroundcolor=\color{backcolour},   
    commentstyle=\color{codegreen},
    keywordstyle=\color{magenta},
    numberstyle=\tiny\color{codegray},
    stringstyle=\color{codepurple},
    basicstyle=\ttfamily\footnotesize,
    breakatwhitespace=false,         
    breaklines=true,                 
    captionpos=b,                    
    keepspaces=true,                 
    numbers=left,                    
    numbersep=5pt,                  
    showspaces=false,                
    showstringspaces=false,
    showtabs=false,                  
    tabsize=2
}

\lstdefinestyle{json}{
  basicstyle=\normalfont\ttfamily,
  numbers=left,
  numberstyle=\scriptsize,
  stepnumber=1,
  numbersep=8pt,
  showstringspaces=false,
  breaklines=true,
  frame=lines,
  backgroundcolor=\color{white},
  stringstyle=\color{blue},
  literate=
   *{0}{{{\color{blue}0}}}{1}
    {1}{{{\color{blue}1}}}{1}
    {2}{{{\color{blue}2}}}{1}
    {3}{{{\color{blue}3}}}{1}
    {4}{{{\color{blue}4}}}{1}
    {5}{{{\color{blue}5}}}{1}
    {6}{{{\color{blue}6}}}{1}
    {7}{{{\color{blue}7}}}{1}
    {8}{{{\color{blue}8}}}{1}
    {9}{{{\color{blue}9}}}{1}
}

\author{%
Joseph Gesnouin$^{1}$ \quad Yannis Tannier$^{1}$ \quad Christophe Gomes Da Silva$^{1}$ \quad Hatim Tapory$^{1}$ \\
\textbf{\quad Camille Brier$^{1}$ \quad Hugo Simon$^{1}$ \quad Raphaël Rozenberg$^{1}$}\\
\textbf{\quad Hermann Woehrel$^{1}$ \quad Mehdi El Yakaabi$^{1}$ \quad Thomas Binder$^{1}$}\\~\\
\textbf{\quad Guillaume Marie$^{1}$ \quad Emilie Caron$^{1}$  \quad Mathile Nogueira$^{1}$ \quad Thomas Fontas$^{1}$ } \\
\textbf{\quad Laure Puydebois$^{1}$ \quad Marie Theophile$^{1}$ \quad Stéphane Morandi$^{1}$ \quad Maël Petit$^{1}$} \\
\textbf{\quad David Creissac$^{1}$ \quad Pauline Ennouchy$^{1}$ \quad Elise Valetoux$^{1}$ } \\~\\
\textbf{Céline Visade}$^2$ \quad \textbf{Severine Balloux}$^2$  \quad \textbf{Emmanuel Cortes$^{2}$} \\
\textbf{Pierre-Etienne Devineau}$^3$ \quad \textbf{Ulrich Tan}$^3$ \quad \textbf{Esther Mac Namara}$^1$  \quad \textbf{Su Yang$^{1}$} \\ \\
$^1$ French Ministry of Economics and Finance - Directorate General of Public Finances \\
$^2$General Secretariat of the French Government - Legal and Administrative Information Directorate \\
$^3$French Interministerial Digital Department \\
\texttt{$^1$ name.fname@dgfip.finances.gouv.fr} \\
\texttt{$^2$ name.fname@dila.gouv.fr} \\
\texttt{$^3$ name.fname@data.gouv.fr}
}

\begin{document}

\maketitle

\begin{figure}[H]
  \centering
    \includegraphics[scale=0.25]{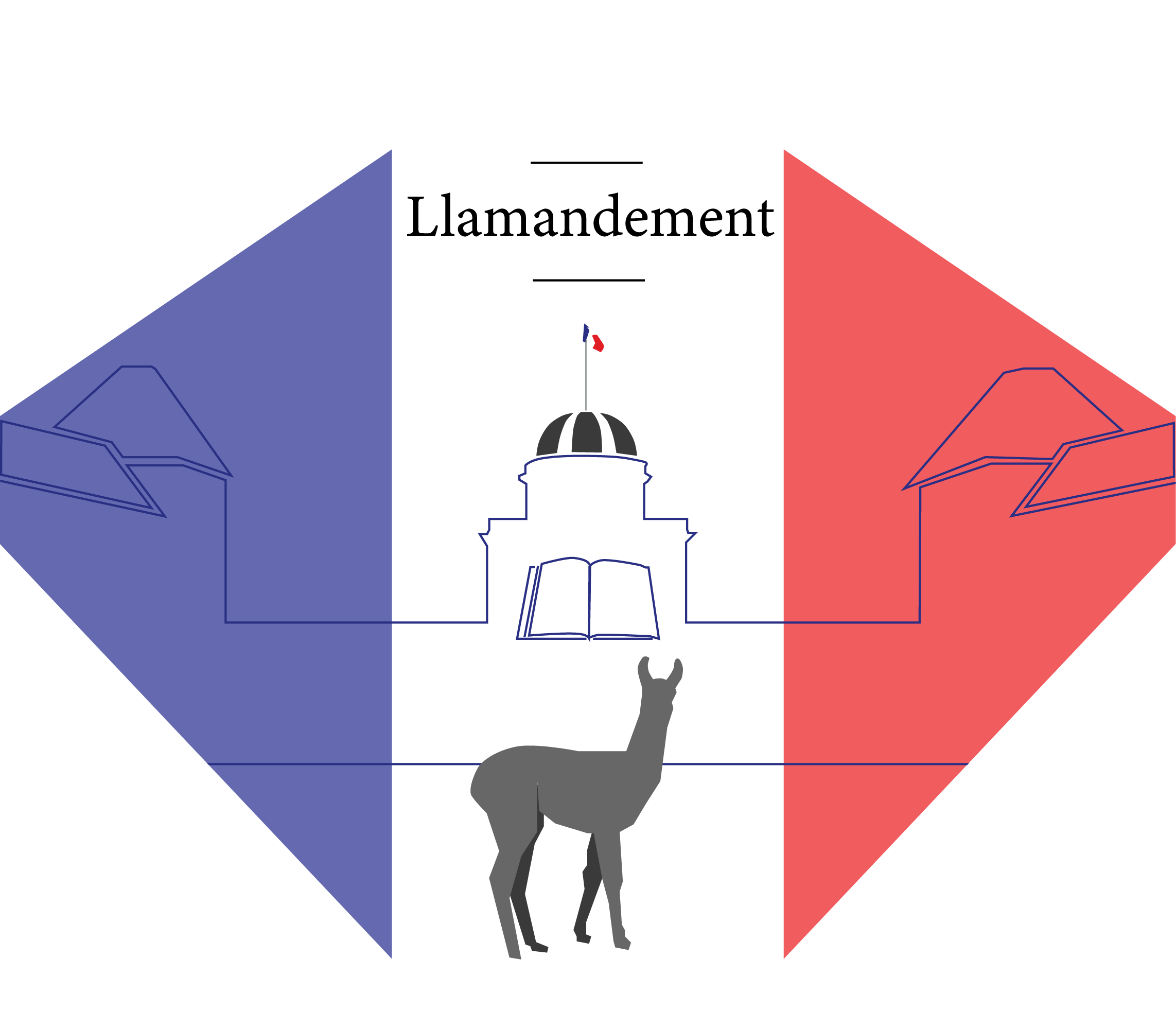}
\end{figure}

\begin{abstract}
This report introduces \textbf{LLaMandement}, a state-of-the-art Large Language Model, fine-tuned by the French government and designed to enhance the efficiency and efficacy of processing parliamentary sessions (including the production of bench memoranda and documents required for interministerial meetings) by generating neutral summaries of legislative proposals. Addressing the administrative challenges of manually processing a growing volume of legislative amendments, \textbf{LLaMandement} stands as a significant legal technological milestone, providing a solution that exceeds the scalability of traditional human efforts while matching the robustness of a specialized legal drafter. We release all our fine-tuned models\footnote{\url{https://huggingface.co/ActeurPublic/LlaMAndement-13b}} and training data\footnote{\url{https://gitlab.adullact.net/dgfip/projets-ia/LLaMandement}} to the community.
\end{abstract}

\section{Introduction}

\subsection{Generative Artificial Intelligence in the Service of the State}
The State is founded on a collection of principles and values that the administration, through its actions, must enliven and ensure continuity. The smooth operation of institutions requires intense and meticulous work from the public servants tasked with this responsibility.

The use of artificial intelligence should enable the State to better meet the expectations and needs of its citizens, optimizing the burden on its agents to fulfill the public service missions entrusted to them. This could manifest as improved responsiveness to public inquiries about administrative procedures, enhanced efficiency in handling time-consuming and repetitive operations for services, or even in improving the democratic processes that invigorate our institutions.
However, at every turn, the development and use of artificial intelligence by the administration must necessarily adhere to imperatives of transparency and accountability. Openly publishing the methods used to develop and train AI models allows for demonstrating adherence to ethical standards (e.g., encountered biases, inconsistent results, usage limitations, etc.) and legal requirements (e.g., respect for personal data protection, professional secrets, etc.).

The french Digital Republic Act (“Loi pour une République numérique”, \cite{france2016republiquenumerique}) aimed to create an environment conducive to the development of public artificial intelligence projects, notably by establishing the principle of default openness of public data (Article 6) and encouraging the development of Open Source (Article 9). The development and widest possible implementation of open public models are in line with this foundational law.

Within the recent successive revolutions of artificial intelligence, Large Language Models (LLMs) hold a unique place. They have enhanced machines' ability to perform more complex language processing tasks, such as understanding long texts, generating coherent and contextually relevant summaries automatically, or interpreting subtle nuances in language. They can be adjusted or fine-tuned for specific applications, allowing customization according to user needs. This makes them useful in specialized fields like law, medicine, or education, and administration, with its expert procedures and complex use cases, is replete with beneficial applications of this technology.

The process of lawmaking offers striking examples of indispensable procedures for the democratic functioning of the nation but disproportionately occupies the time and attention of public servants.

\subsection{Challenges and Dynamics of Amendment Procedures in French Parliamentary Practice}
The drafting of laws is one of the most significant events in democratic life. This moment, expressing both the aspirations and needs of a society, profoundly and durably influences the lives of citizens. Arising from the legislative power, the construction of laws generally falls within the purview of a parliament, animating the parliamentary life of a country.

The French Constitution\cite{french_constitution1958} defines the nature, role, and powers of the Parliament in France. It shares legislative initiative (\textit{i.e.} the power to propose laws) with the Government. From its drafting to its implementation, a bill undergoes several stages, defined in particular by the internal rules of each of the two chambers composing the Parliament (the National Assembly\cite{rules_nationalassembly} and the Senate\cite{rules_senate}). This legislative procedure can be summarized in five main phases:
\begin{enumerate}
    \item The submission of the bill to one of the two chambers.
    \item The examination of this bill within the chamber to which the text was submitted, usually resulting in a new amended text.
    \item The Shuttle (this new text is transmitted to the second chamber where it is examined and a new version is generally adopted; this new version is sent back to the first chamber where it is examined and a new version is adopted; and so on).
    \item The final adoption of the bill when both chambers eventually adopt an identical text.
    \item The promulgation of the text by the President of the Republic.
\end{enumerate}

The examination of the bill is a crucial moment in the legislative procedure. It involves a debate and a vote on amendments (modifications to the text proposed by Parliament or the Government). Subject to certain restrictions (imposed notably by the Constitution, organic laws, and the rules of the National Assembly and the Senate), this right to amend is unlimited.

Each of the amendments submitted for examination in one of the two chambers is analyzed and processed by agents of the administration working for the government, facilitating the smooth progress of discussions, debates, and votes in Parliament (thanks to the creation of bench memoranda for ministers who will intervene during parliamentary sessions or preparatory work before inter-ministerial meetings, for example). Part of the work in processing these amendments can be summarized as follows:
\begin{enumerate}
    \item The submitted amendments are collected by administration agents immediately after the submission deadline.
    \item Any texts similar to these amendments are identified (notably to ensure a certain coherence in the government's responses to these amendments).
    \item The amendments are then assigned to the various ministries concerned with their content, with the aim of obtaining insights and opinions on these texts.
\end{enumerate}

Since 2020, the Interministerial Digital Management System for Legislative Amendments (SIGNALE\cite{dila:adm-01859732}) has enabled administrative agents to coordinate and work more efficiently on their tasks related to bills (especially amendments). For instance, it facilitates the drafting of bench memoranda and the creation of tables for the preparation of inter-ministerial meetings concerning legislative texts. This platform serves as a suitable support where various tools to optimize the work of administrative agents on bills could be implemented.

\begin{figure}[H]
   \includegraphics[width=\textwidth]{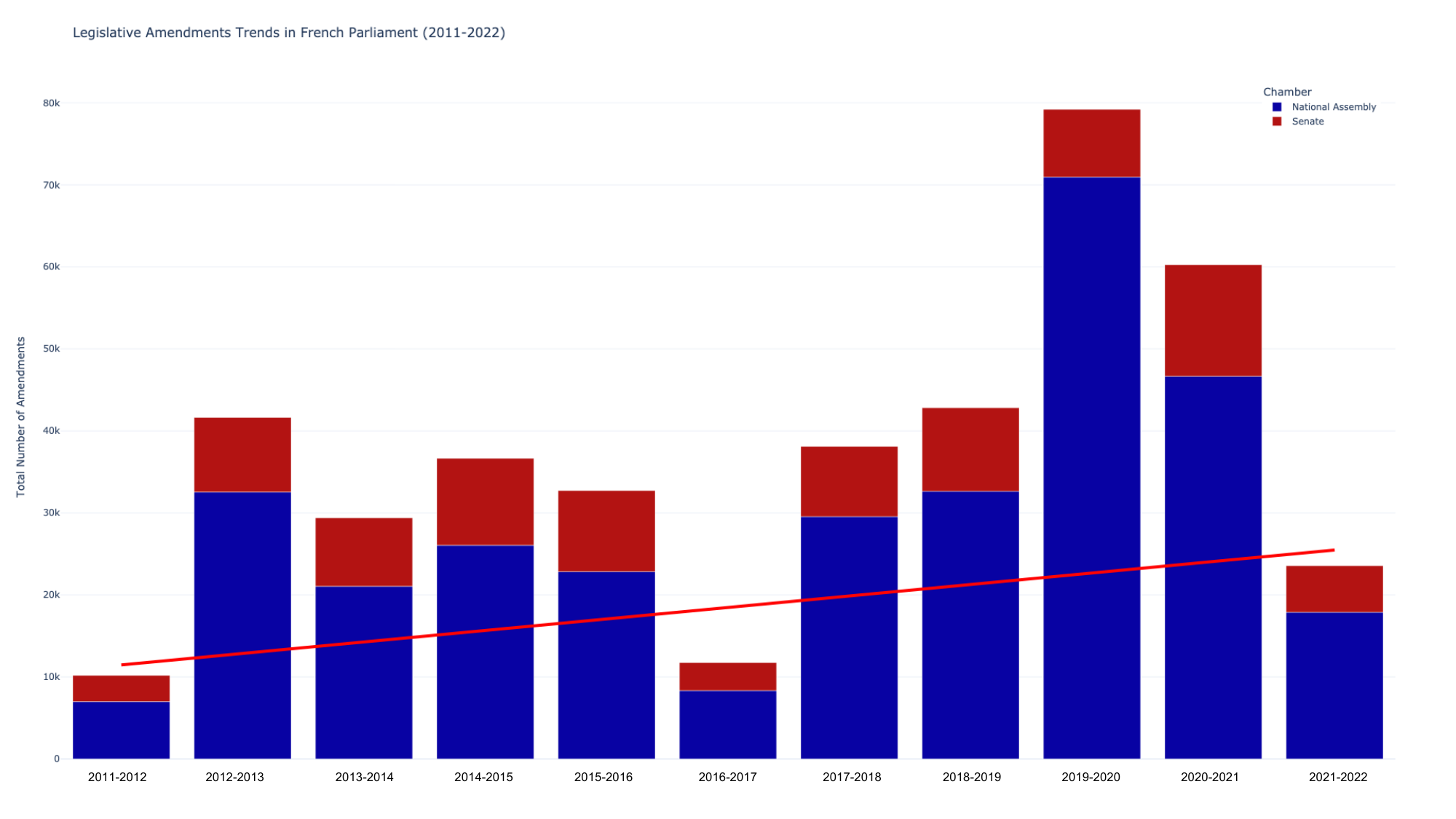}
  \caption{Stacked bar chart displaying the aggregate amendments filed in the French Parliament from 2011 to 2022, segmented by originating chamber. The Senate's contributions are shown in red and the Assemblée Nationale's in blue. A red trend line highlights the overall direction of amendment activity throughout the last decade.}
  \label{fig:barchart}
\end{figure}

The processing of amendments demands significant reliability and precision; it directly contributes to lawmaking. It must adhere to various standards, including the internal rules of the National Assembly and the Senate, which impose a certain timeframe\cite{rules_nationalassembly} for processing amendments. The number of amendments also increases over time as shown in Fig\ref{fig:barchart}, often reaching thousands for a given bill. All these administrative constraints weigh on the work of administration agents.

Beyond its purely procedural and administrative conception, lawmaking also represents a significant political moment: it allows parties to build their strategy and leave their mark. The right to amend is thus a political instrument: it can, for example, transform into a tool for legislative obstruction (an extremely high number of amendments can be proposed to significantly slow down the examination of bills) in service of a political maneuver\cite{obstructionlegislative}, and the content of the amendment itself can highlight a partisan orientation. This instrumentalization directly affects the missions of agents who are obliged to provide impartial work. It also contributes to the inflation of the number of amendments (even as the constraints remain the same), further limiting the maximum duration of their management, thus affecting the efficiency, precision, and reliability of the analyses necessary for lawmaking.

The automation of part of the amendment processing work (assignment, identification of similar texts, summarization) envisioned within the \textbf{LLaMandement} project aims to facilitate and improve the quality of the work of administration agents by alleviating the time constraint. It also ensures a certain level of reliability and precision, providing neutral results. Serving as an assistive tool, it supports administrative agents in efficiently navigating the complexities of bill analysis and amendment processes. It aids in the preparation of documents such as bench memoranda and facilitates the coordination of interministerial meetings (RIMs), particularly in periods of heightened legislative activity. This utility is crucial in ensuring a streamlined and objective approach to legislative review and documentation.

\subsection{The Practical Exercise of the Right of Amendment: An Administrative and AI Perspective}
\begin{figure}[!h]
   \includegraphics[width=\textwidth]{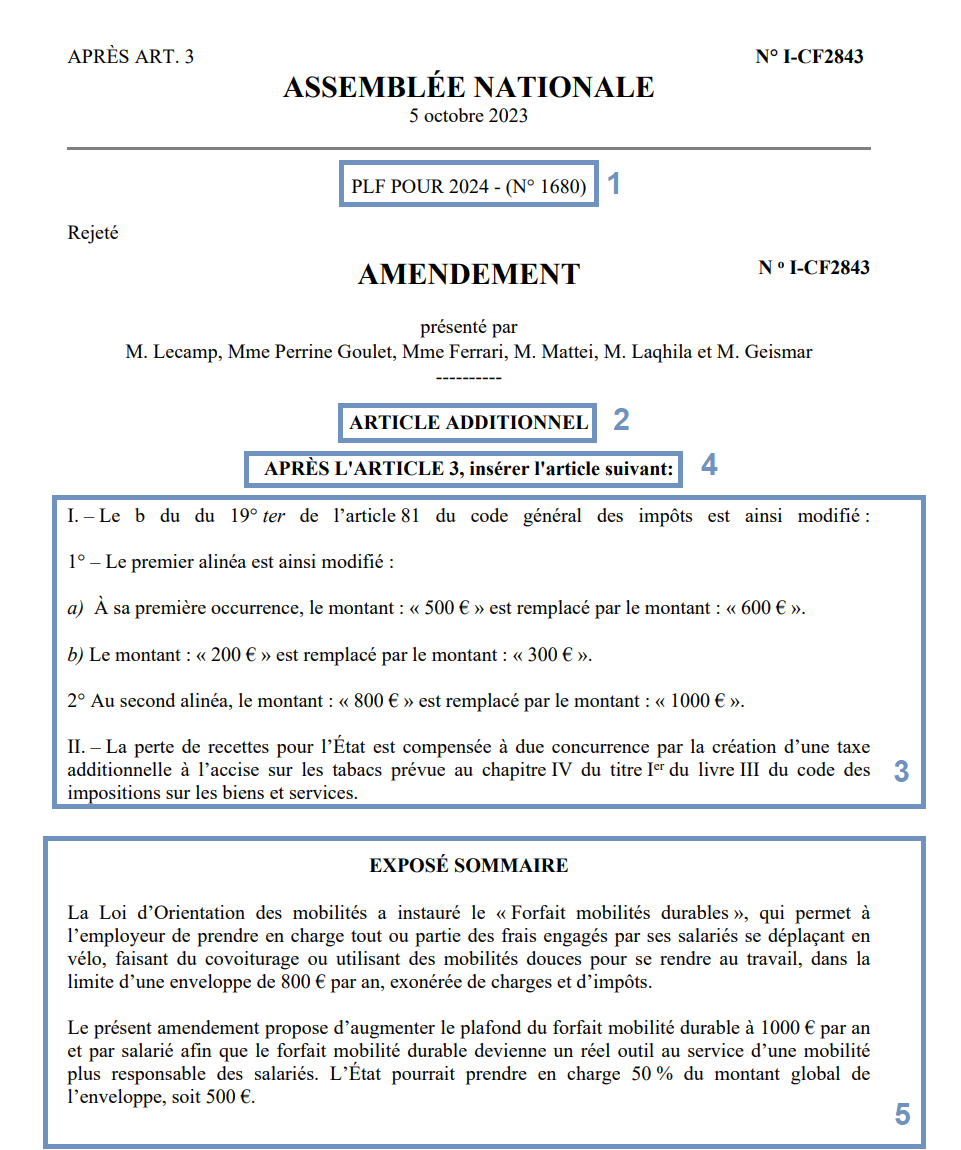}
  \caption{The image below shows an amendment in its most formal form as it should be submitted by any Member of the french Parliament. \textbf{(1)} At the very top, the document indicates the legislative bill to which the amendment is to be applied, providing context and reference for the proposed changes. \textbf{(2)} The amendment is identified either as targeting a specific existing article or as an 'Article additionnel' if it introduces new provisions not covered by existing articles. \textbf{(3)} The main body of the amendment follows, detailing the exact legislative changes proposed, known as the dispositif. This is where the parliamentarian's proposed text to be inserted into or to modify the bill is articulated. \textbf{(4)} The dispositif begins with a header that precisely locates where within the bill the amendment will take effect, such as specifying after which article the new content is to be inserted. \textbf{(5)} Finally, every amendment is accompanied by a rationale, outlined in the 'EXPOSÉ SOMMAIRE,' which provides the reasons and objectives behind the amendment, a mandatory element for the amendment's admissibility.}
  \label{fig:amendement}
\end{figure}

In the practical implementation of the amendment right, legislative departments across various French ministries delegate three essential tasks to their coordinating bureau: amendment attribution, similarity research and amendment summaries. These tasks are crucial in ensuring the efficient processing and integration of amendments into the legislative framework, in particular through the drafting of bench memoranda. Nevertheless, the efficiency with which they are carried out, as well as the reliability and accuracy in some cases, can be greatly enhanced by automation techniques.

\subsubsection{Amendment attribution and similarity research: using simple language processing techniques}
Amendment attribution and similarity research are tasks for which simple tools such as expert systems or fuzzy matching can considerably improve efficiency, reliability, and accuracy of execution (particularly in the case of assigning amendments). These methods, which have been tried and tested in a variety of applications for several decades, do not require advanced machine learning techniques and do not represent any particular technological challenge, even though they can significantly enhance the performance of administrative agents in their amendment processing tasks. The automatic assignment of amendments and the automatic search for similarities (projects related to \textbf{LLaMandement}), which are integral parts of the processing of bills in the legislative procedure, will therefore be described only briefly below in this report.
\begin{itemize}
    \item \textbf{Amendment Attribution}: The foundational step in this process involves identifying key terms within legislative texts. Typically, ministries are structured to facilitate the prompt distribution of amendments to appropriate departments or personnel. This conventional method, often time-consumming and error-prone, can be greatly enhanced through the use of natural language processing techniques for keyword analysis. These tools expedite the classification and distribution of amendments by analyzing their content, as represented by items \textbf{(3)} and \textbf{(4)} in Figure \ref{fig:amendement} and determining the most relevant departmental match. Due to the numerous rules defining the structure of bills, the tools used are more akin to simple expert systems, not requiring advanced machine learning techniques. This highlights the need for systems that are versatile enough to adapt to the distinct procedural nuances of each ministry, making the widespread adoption of this amendment allocation process intrinsically complex at an interministerial level.
    \begin{itemize}
        \item \textbf{Amendment Attribution at the French Directorate General of Public Finances}: The use of AI tools in the Tax Legislation Directorate for the 2024 finance bill has been notably successful, demonstrating their efficacy in specific legislative contexts. In this implementation, the tools achieved an 94\% auto-attribution rate for roughly 5400 amendments in under 10 minutes, showcasing their potential for efficiency and accuracy. Yet, the replication of this NLP-driven process across various ministries is challenging due to their diverse practices and operational methods.
    \end{itemize}
    \item \textbf{Similarity Research}: A critical aspect of the legislative process consists of identifying similarities in amendments and legislative texts. This task extends beyond the current bill under consideration, incorporating a comprehensive review of historical amendments. The primary objective is to ensure each amendment's uniqueness and prevent redundancy. This practice is vital in legislative processes as it avoids duplicative efforts, allowing agents to efficiently leverage previously written, attributed, and summarized work for similar past amendments. This procedure entails a considerable constraint, demanding the engagement of specialists with comprehensive knowledge of their legislative area. The use of simple AI tools is critical in enhancing the efficiency of similarity research in legislative contexts, especially given the standardized structure and vocabulary of amendments. Employing learning-free methodologies, such as fuzzy matching \cite{max_bachmann_2021_5584996} and Jaro-Winkler distance \cite{jaro1989advances,winkler1999state}, these tools excel at swiftly scanning and comparing large legislative datasets. Their effectiveness is primarily attributed to the amendments' uniform structure, focusing the analysis on their main body, as represented by item \textbf{(3)} in Figure \ref{fig:amendement}. This approach, highly scalable and adaptable to various volumes of amendments, circumvents the need for complex models such as LLMs \cite{le2019flaubert,wang2022text,reimers2019sentence,martin2020camembert}.
    \begin{itemize}
    \item \textbf{Similarity Research at the French Directorate General of Public Finances}: A practical demonstration of this has been done during the first reading of the 2024 finance bill, which contained roughly 5400 amendments. In this instance, fuzzy matching tools successfully identified approximately one-sixth of the amendments as having existing counterparts, doing so without necessitating domain-specific expertise. This capability is extremely valuable, as it allows legislative staff to efficiently refer to and build upon work from similar previous amendments, thereby markedly enhancing the efficiency of the legislative review process.
\end{itemize}
\end{itemize}

\subsubsection{Amendements summaries: using Large Language Models (LLMs)}
   The final task of preparing a concise summary holds immense significance in the legislative process. Given that an average amendment spans roughly two pages and with tens of thousands of amendments processed annually, the ability to quickly grasp the essence of each amendment is crucial. Summaries become indispensable tools for various stakeholders - including the government, ministers, commission members, deputies, senators, administrative agents who monitor and manage the legislative process, journalists or even citizens. They need to rapidly exchange and discuss the content of an amendment without the necessity of delving back into the full text. This is where AI-driven synthesis tools can prove useful. Contrasting with the specialized expertise needed for automatic amendment attribution or the simpler use of fuzzy matching for similarity analysis, AI-driven summarization, adaptable for use in various ministerial contexts, requires more advanced methods, like large language models (LLMs). By generating precise, neutral summaries that distill the essence of these complex legal texts, they make the information easily digestible and accessible, facilitating efficient communication and informed decision-making among all involved parties.
   
\textbf{Summaries at the French Directorate General of Public Finances}: AI's successful application for summarization during the 2024 finance bill highlighted its potential for broad application across different ministries.
\\~\\
Given the technical complexity of the operation and development of LLMs (compared with expert systems or fuzzy matching used in amendment allocation and similarity search), which are necessary for the automatic drafting of neutral, reliable and accurate summaries in an efficient way within the framework of the legislative procedure, the remainder of this report will focus on the technical aspects and results obtained using LLMs for amendment summaries. Indeed, these tools need a detailed explanation, including the specifics of the models used, how they are trained, and the overall strategy for their implementation.
Considering the diverse nature of legislative bills and their unique implications, this technical report aims to showcase the enhancements made to a standard large language model (\textit{i.e.} \textbf{LLaMA} \cite{touvron2023llama} into \textbf{LLaMandement}) for its effective use in various ministerial contexts, in addition to drafting bench memoranda. We delve into the technical adjustments and strategic implementations undertaken to adapt \textbf{LLaMandement} for summarizing a wide array of french legislative texts. This effort demonstrates our commitment to fine-tuning LLMs to meet the specific requirements of different ministries.
\begin{mybox}
\textbf{Enhancing Legislative Efficiency: AI Applications} \\~\\
In legislative administration, the integration of Artificial Intelligence and automation tools, is speeding up the revision of legislative measures and the drafting of documents such as bench momranda and preparatory work for RIMs. This involves streamlining three critical operations: \textbf{(1) Amendment Attribution}, where AI facilitates rapid and accurate distribution of amendments to pertinent departments. \textbf{(2) Similarity Detection}, utilizing techniques like fuzzy matching to identify and avoid redundant amendments efficiently. \textbf{(3) Amendment Summarization}, where AI-generated concise summaries support stakeholders in swift and informed decision-making. 
\\~\\
While the former tasks may be more routine, the latter—\textit{summarization}—necessitates a sophisticated, calibrated approach. The evolution of \textbf{LLaMA} \cite{touvron2023llama} into the specialized \textbf{LLaMandement} model reflects the focused effort to satisfy the nuanced demands of legislative text processing.

This report delves into the strategic employment of LLMs, underpinning the efficacy of \textbf{LLaMandement} in delivering impactful summaries across the diverse landscape of government sectors.
\end{mybox}

\section{Related Work: Technical Solutions Through Large Language Models (LLMs)}

The increasing complexity and volume of amendment proposals, coupled with the detailed task of drafting bench memoranda, present formidable challenges that necessitate innovative technological interventions. In response, Large Language Models (LLMs) have emerged as a plausible solution, with their advanced processing abilities and scalable nature. This section delves into the development, evolution, and capabilities of LLMs, emphasizing their usefulness in simplifying and improving the French legislative process.

\subsection{The Rise of Large Language Models in AI: A Paradigm Shift}
Large Language Models (LLMs) signify a major breakthrough in the field of artificial intelligence, particularly transforming the landscape of language processing. Key examples of these models include, GPT-3\cite{brown2020language}, PaLM\cite{chowdhery2022palm}, \textbf{LLaMA}\cite{touvron2023llama}, BLOOM \cite{workshop2022bloom}, Falcon\cite{refinedweb} and Mistral\cite{jiang2023mistral}. These models demonstrate exceptional capabilities in various language tasks, a result of their scalability in terms of model size and data handling \cite{kaplan2020scaling}. As they expand, they exhibit new abilities, surpassing the limitations of smaller-scale models like BERT \cite{devlin2018bert}.

A major advancement in LLMs is their capability for in-context learning, allowing them to interpret and respond appropriately within the contextual framework of a text. This bypasses the necessity for explicit scenario programming. Highlighted in studies like Brown \textit{et al.} \cite{brown2020language}, this shift represents a departure from traditional models reliant on extensive training with large datasets \cite{wang2023learning}. Such development in LLMs enhances their versatility, enabling applications in language understanding, reasoning, planning, and applying common sense, thereby fostering a more natural interaction with language.

\subsection{Beyond Language: Separating Substance from Political Messaging}
The evolution of LLMs has broadened their scope, moving them from conventional language tasks to sophisticated analysis in complex domains. A key advancement is their ability to discern substantive content from stylistic or rhetorical elements, ensuring a focus on core information rather than presentation style. This distinction is crucial, especially in legislative analysis where maintaining objectivity is of utmost importance. Legislative proposals often embed biases, both in their content and presentation, which can subtly influence how the information is framed and perceived. In democratic environments, it is imperative to clearly separate the essential content of these proposals from their political messaging. This separation is vital to avoid skewed public perception and biased political discourse. It underscores the importance of having tools capable of discerning the fundamental aspects of legislative proposals from their persuasive elements.

A notable study illustrating this expanded capability is the work by Wu \textit{et al.}\cite{wu2023large}. This research delves into the application of LLMs for identifying latent ideologies of lawmakers. Through their analysis, LLMs demonstrate an ability to discern implicit biases or ideological undercurrents that may not be overtly evident in the text. This ability to interpret subtle nuances and underlying tones in texts represents a significant enhancement over traditional text analysis methods.

The Med-PaLM project\cite{singhal2023large,singhal2023towards} offers another perspective, highlighting the ability of LLMs to provide factual, unbiased responses in medical scenarios. It aims to utilize LLMs' processing power to navigate vast medical information and produce responses that are not only accurate but also devoid of prejudicial inclinations. This demonstrates the potential of LLMs as reliable sources of information in fields where accuracy is critical.

LLMs' proficiency in sentiment analysis is essential for isolating factual content from emotional or biased language, particularly in legislative contexts. This skill ensures that legislative discussions and decisions are based on the actual substance of proposals, rather than their rhetorical framing. The continued advancement of LLMs enhances their role in facilitating transparent and unbiased legislative processes, cementing their importance not only as language processing tools but as critical instruments for nuanced analysis in various domains, thereby aiding informed decision-making and policy formulation.

\subsection{Challenges in Domain-Specific Applications}
While LLMs have demonstrated remarkable proficiency in general language tasks, their application in specialized domains such as law \cite{blair2023can,trautmann2022legal,choi2023chatgpt}, medicine\cite{nov2023putting,yang2023evaluations}, finance \cite{shah2023zero,araci2019finbert,son2023beyond}, and scientific research\cite{jin2019pubmedqa,krithara2023bioasq}presents unique challenges. These challenges primarily stem from deficiencies in domain-specific knowledge, the capacity to effectively leverage such knowledge, and the risk of generating factually inaccurate content.

For instance, in the legal sector, the fine-grained understanding required for complex legal language processing poses significant challenges for LLMs. Studies such as Huang et al. \cite{huang2023lawyer} illustrate the struggles of LLMs in accurately distinguishing and interpreting legal terminologies, such as the difference between 'deposit' and 'down payment' in Chinese contract law.
In healthcare, where the stakes are high, the accuracy of information is critical. Wang et al. \cite{wang2023survey} highlight the risks associated with misinformation in medical advice generated by LLMs, underlining the necessity for high factual accuracy and reliability.

To address these challenges, there is a growing consensus in the academic community about the need for more specialized training and data incorporation. The development of domain-specific LLMs, such as Med-PaLM\cite{singhal2023large,singhal2023towards} for healthcare or BloombergGPT\cite{wu2023bloomberggpt} for finance, represents a step towards this goal. These tailored models demonstrate improved performance in their respective fields by leveraging domain-specific knowledge bases and training datasets.
\begin{mybox}
\textbf{Adapting Large Language Models: From Broad Applications to Specialized French Legislative Analysis}\\~\\
LLMs have demonstrated remarkable efficiency in general language processing, showcasing a robust ability to synthesize and articulate complex ideas. A notable advancement in their application is the capacity to separate substantive content from rhetoric, an essential feature for \textbf{LLaMandement}. However, when it comes to domain-specific tasks, LLMs encounter challenges due to the need for specialized knowledge and nuanced understanding. Addressing these limitations is crucial for \textbf{LLaMandement}, where fine-tuning LLMs to grasp the unique intricacies and legislative context becomes indispensable.
\end{mybox}


\section{Materials and Methods}
This section outlines the methods and materials used, detailing the model selection process, fine-tuning techniques, and data collection strategies that contributed to the model's capability to process and understand french legislative texts with high accuracy and efficiency.

\subsection{Model Selection}
The development of \textbf{LLaMandement} hinged on a critical objective: to achieve a level of performance in processing and understanding legislative texts that is sufficiently good from a business perspective using the simplest yet most effective computational models. This pursuit led us through an exploration through various model complexities, starting with classical transformer models \cite{raffel2020exploring,xue2020mt5} and progressing to Large Language Models (LLMs)\cite{touvron2023llama,refinedweb,jiang2023mistral}. Our guiding principle was not merely to chase the latest advancements but to discern the most suitable model that could seamlessly deal with the intricacies of the french legislative language.

To this end, the \textbf{LLaMA} 70B \cite{touvron2023llama} model emerged as a pivotal tool. Its versatility was tested across different learning paradigms:

\begin{itemize}
    \item \textbf{Zero-shot learning}: We evaluated the model's inherent ability to interpret and summarize french legislative amendments. This phase was crucial in assessing the natural linguistic and analytical prowess of the model. A typical zero-shot prompt was:
    \begin{lstlisting}[language=Python]
    prompt_zero_shot = "Here's an amendment: "+text+"
    Could you summarise this amendment for me in one sentence, as if I were a lawyer?"\end{lstlisting}
    \item \textbf{Few-shot learning}: We  gained a glimpse into the model's adaptability. Here, the \textbf{LLaMA} 70B was provided with minimal examples, simulating the real-world scenario where legislative texts often present unique, unprecedented cases. This phase tested the model’s efficiency in quickly assimilating new information and producing coherent, legally accurate summaries. A typical few-shot prompt was:
    \begin{lstlisting}[language=Python]
    prompt_few_shot = "You are a legal professional tasked with summarizing an amendment. Your task is to summarize the amendment in one sentence maximum. Here are some examples of amendment summaries made by lawyers:
    
        -Implement an exceptional mechanism for unlocking employee savings (profit-sharing and participation) up to 30,000 euros for funds placed in a savings plan before January 1, 2024, at the employee's request between July 1 and December 31, 2023.
        -Increase to 50% the exemption from TFPNB (Tax on Non-Built Properties) applicable to agricultural land.
        -Restore the arrangement provided in Article 199 terdecies-0 B of the CGI (General Tax Code) for loans taken out between the date of promulgation of the present law and December 31, 2025.
        -Exempt from VAT non-profit participatory housing projects intended for primary residences.
        -Ensure the application of the reduced VAT rate of 5.5% to energy renovation works that have already been subject to a quote and a deposit payment before the publication of the order modifying the scope of eligible works.
        -Exempt from the housing tax (TH) social and medico-social establishments and services (EMS) and private health establishments of collective interest.
        -Abolish the real estate wealth tax (IFI).
    
    You should draw inspiration from these summaries for the next amendment. The summary should be neutral, start with an infinitive verb, maintain legal vocabulary, and include important figures. Here's an amendment: "+text\end{lstlisting}
\end{itemize}

These experiments were instrumental in pushing the \textbf{LLaMA} 70B to its limits, thereby revealing the 'glass ceiling' of current LLMs in dealing with complex french legislative content. This understanding was vital for setting a benchmark that informed our fine-tuning strategy.  As we transitioned to the fine-tuning phase, a strategic shift occurred. While the initial stages with the \textbf{LLaMA} 70B model provided us with insights into the upper performance limits, we then opted for a more balanced \textbf{LLaMA} model tailored for fine-tuning. This choice was dictated by the need to balance computational power with efficiency and sustainable resource usage, especially considering the operational demands of legislative environments.



\begin{mybox}
    \textbf{Balancing Efficiency with Responsible Resource Usage}\\~\\
    Our selection process was designed to first identify the performance limits of the most advanced LLMs. This strategy ensured an informed decision for fine-tuning, aligning our selection with the dual objectives of high efficiency and sustainable resource usage in legislative text processing.
\end{mybox}

\subsection{Fine-Tuning: Low Rank Adaptation}

Following this, we implemented Low-Rank Adaptation (LORA \cite{hu2021lora,dettmers2023qlora})\footnote{https://github.com/lm-sys/FastChat/} for fine-tuning our selected \textbf{LLaMA} model. LORA is recognized for its efficiency in fine-tuning deep neural networks with minimal additional computational burden. This technique introduces adaptability in model training, especially beneficial for extensive models like \textbf{LLaMA}, by inserting additional parameters at a low rank. These parameters are specifically designed to adapt the model's responses to the intricacies of legislative language without extensive retraining or significant alterations to the model's structure.

In our application, the settings for LORA were calibrated as follows:
\begin{itemize}
    \item \textbf{Learning Rate (LR)}: We set the LORA learning rate, $\sigma$ = 2e-5, lower than typical fine-tuning rates to allow for gradual, stable adaptation of the model.
    \item \textbf{Adaptation Depth ($lora\_r$)}: We set its value to 64. This parameter specifies the rank (dimension) of the low-rank matrix in LoRA. In our \textbf{LLaMA} 13B model, fine-tuning with LORA affected approximately 0.40\% of the weights (\textit{i.e.} 50m parameters) . This balance is crucial for achieving adaptability to new tasks or data while retaining the original design and knowledge of the pre-trained model.
    \item \textbf{Decay Rate}: A decay rate of 0.01 was employed for regularization, reducing the risk of overfitting to specific legislative text structures.
    \item \textbf{LORA Alpha ($\alpha$)}: Set to $\alpha$ = $16$, this parameter controls the scaling of the LORA adjustments. This parameter is key in fine-tuning how the LoRA modifications affect the model, allowing for more precise tuning of the model's responses to the nuances in legislative texts.
    \item \textbf{LORA Dropout}: A dropout rate of $0.1$ was applied to the LORA layers to prevent overfitting and enhance generalization capabilities.
    \item \textbf{Optimizer and Scheduler}: A cosine learning rate scheduler with a warmup ratio of 0.03 was utilized to optimize the training process \cite{loshchilov2016sgdr}.
\end{itemize}

\subsection{Data Collection}

For the development of \textbf{LLaMandement}, we sourced our data from SIGNALE\cite{dila:adm-01859732}, a digital platform integral to the French government's legislative process. This platform provided a rich and varied legislative dataset, essential for our model's performance across different ministerial contexts.

\begin{figure}[!ht]
\includegraphics[width=\textwidth]{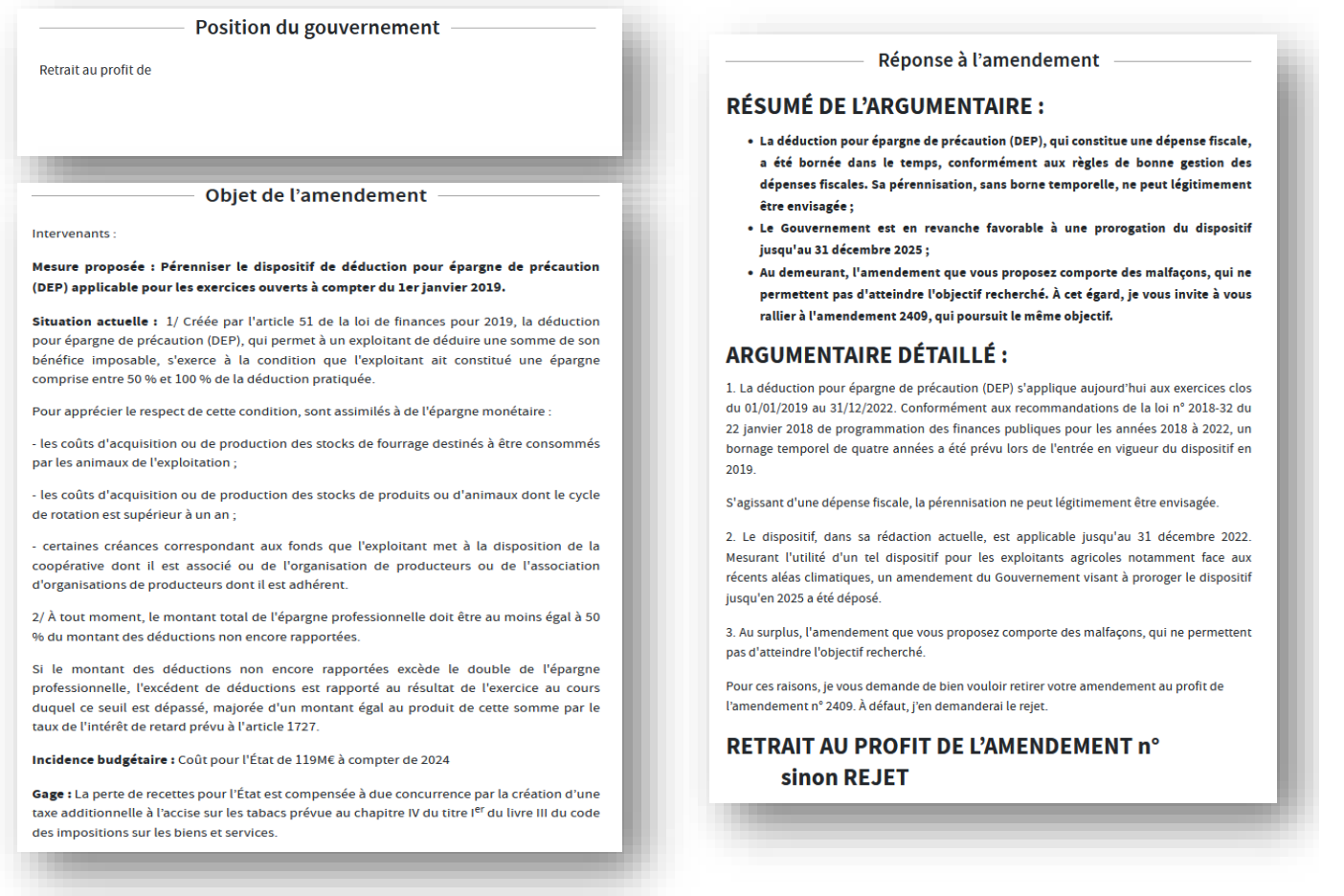}
\caption{A bench memorandum detailing the government's position and a detailed response to an amendment proposal. The left side outlines the current situation, the proposed measure, budgetary impact, and a tax note. The right side provides the government's stance, offering an argument on the proposal's implications, followed by a detailed rationale for supporting or opposing the amendment.}
\label{fig:fiche_de_banc}
\end{figure}

A vital part of our dataset was the ministers' bench memoranda. These documents, used by French ministers during parliamentary sessions, are comprehensive and contain detailed legislative information. Crucially, each file begins with a factual summary of the amendment, prepared by an administrative drafter. These summaries offer an objective, succinct overview, essential for understanding the amendment's content and intent. Illustrating this point, Figure \ref{fig:fiche_de_banc} presents a typical bench memorandum highlighting the structure and type of information typically included in these files.

Building upon the example provided, it is clear that the incorporation of these factual summaries of the proposed measures into our dataset is invaluable. They enable \textbf{LLaMandement} to access clear, condensed versions of complex french legislative texts. Through processing these summaries, the model learns to extract crucial information from detailed legislative documents, an essential skill for proficient legislative analysis.

\begin{figure}[!ht]
\includegraphics[width=\textwidth]{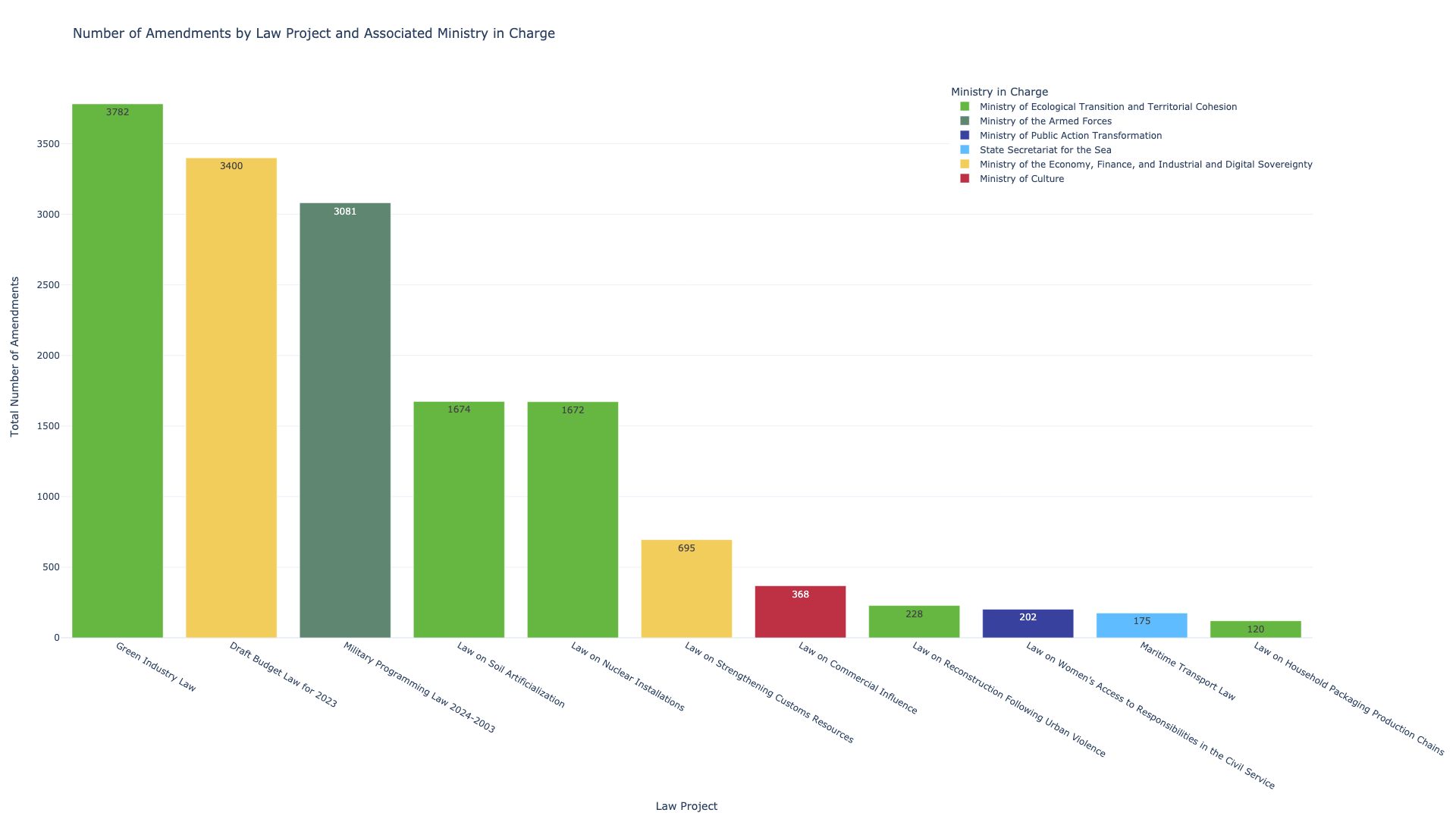} 
\caption{Distribution of amendments and summaries across various 2023 bills in our dataset, showcasing the extensive legislative activity, reflecting \textbf{LLaMandement}'s comprehensive fine-tuning scope.}
\label{fig:dataset}
\end{figure}

Figure \ref{fig:dataset} illustrates the distribution of 15,397 pairs of amendments and their corresponding summaries across various 2023 bills in our training dataset. Following the initial selection, these pairs were further sorted and standardized to exclude any with missing or low-quality summaries. This additional step in our curation process further enhances the quality and relevance of the training material, which we have made publicly accessible\footnote{\url{https://gitlab.adullact.net/dgfip/projets-ia/LLaMandement}}. This meticulous approach to dataset construction reflects the model's extensive and diverse legislative learning base, encompassing:

\begin{itemize}
\item \textbf{The Military Programming Bill 2024-2030}: A comprehensive plan setting out the future of France's defense and security, detailing the allocation of resources and strategic objectives to safeguard national interests.
\item \textbf{The Green Industry Bill}: A forward-looking initiative that encapsulates the government's commitment to ecological transition, charting a path for sustainable development within the industrial sector.
\item \textbf{The Finance Bill 2023}: A fiscal roadmap that delineates the country's economic strategy, shaping taxation and budgetary allocations to steer economic growth and stability.
\item \textbf{Additional significant bills}: These cover a range of critical societal and economic issues, showcasing legislative responses to contemporary challenges. They include measures to curtail land artificialization, thereby preserving natural habitats; proposals to bolster France's energy independence through nuclear innovations; strategies to equip customs with the means to counter emerging security threats; regulations designed to ensure transparency and accountability among social media influencers; initiatives aimed at fostering gender equality and enhancing the representation of women in public service; solutions to the complex problems of urban redevelopment; actions to prevent social dumping in the maritime industry; and reforms to unify and streamline waste management practices for greater efficiency and environmental responsibility.
\end{itemize}

\begin{mybox}
\textbf{Objective of Diverse Data Collection}\\~\\
The varied legislative data collected for \textbf{LLaMandement} is designed to enhance its analytical precision and adaptability across different legislative contexts in french. This diverse dataset equips the model to effectively navigate and interpret the complexities inherent in governmental policy-making and legislative discussions, thus enriching its technical accuracy and deepening its capacity to succinctly summarize multifaceted legislative content.
\end{mybox}

\section{Experiments}
This section presents a detailed examination of the methodologies and outcomes of our experimental analysis, focusing on the effectiveness of various natural language processing models, including the \textbf{LLaMandement}, in summarizing complex legislative texts. Through a combination of quantitative assessments and qualitative insights, we endeavor to showcase the advancements and practical applicability of these models in the specialized field of legislative drafting and summarization.

\subsection{Evaluation Methodology}
To evaluate the effectiveness of different summarization models in our research, we employed a metric designed to align closely with professional human judgment. A selected panel of ten experienced fiscal drafters participated in a blind review process, assessing the quality of summaries produced by various methodologies. Each drafter was tasked with rating the summaries of legislative amendments on a scale of 0 to 10, with 10 being the highest. These individual scores were then averaged and scaled to a 20-point system to provide a standardized measure of performance.

This comprehensive evaluation included comparisons with summaries created by the drafters and their peers over the past year, as well as those generated by our models for 30 specifically selected amendments not included in the 2023 Finance Bill training data. Such an approach ensures a robust and impartial assessment, leveraging the practical knowledge and critical expertise of veteran legislative professionals.

\begin{mybox}
    \textbf{Setting the Benchmark: Fiscal Drafters' Self-Rating as a Model Performance Target}\\~\\
The legal drafters, on average, rated their own summaries at a notable 16.5 ($\pm$ 5.2) out of 20, establishing a benchmark for automated summaries and providing a clear, quantifiable target for the performance of our computational models.
\end{mybox}

\subsection{Quantitative Results}

\begin{figure}[!h]
\includegraphics[width=\textwidth]{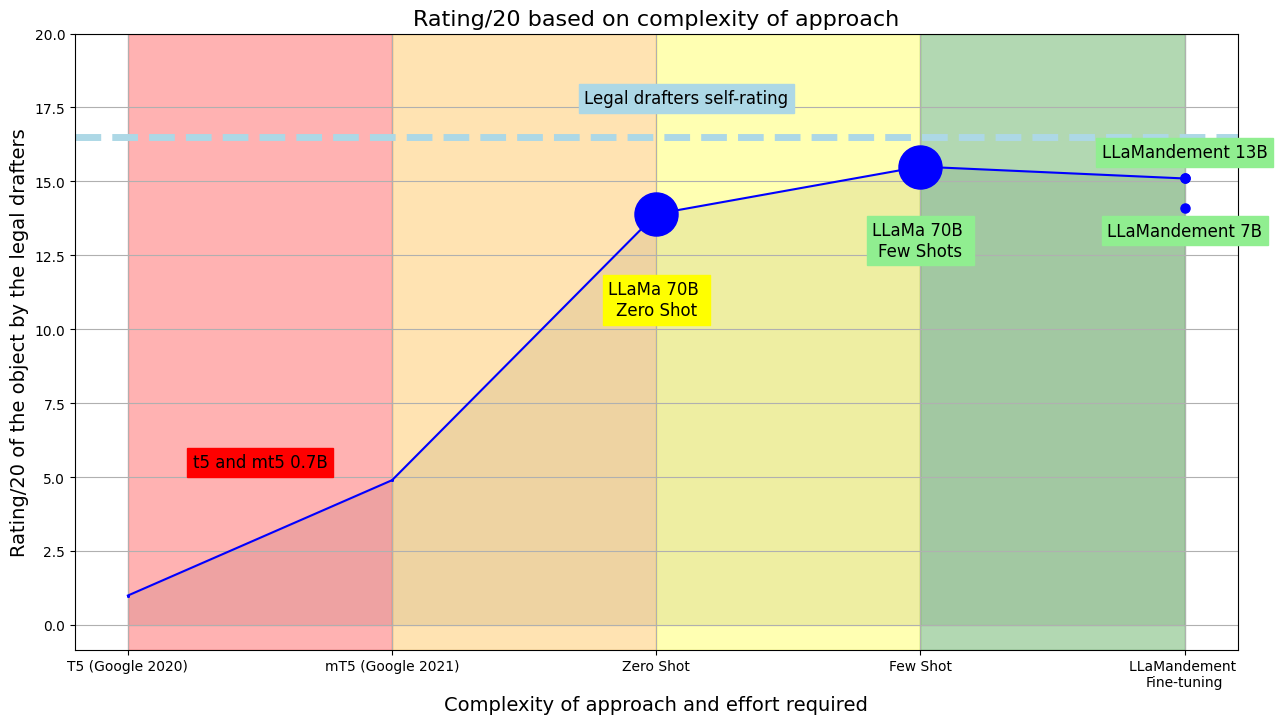}
\caption{Correlation of Model Size and Complexity to Performance in Legislative Text Analysis. The graph depicts the evolution of NLP models' performance scores as judged by legal drafters, with the size of each point correlating to the model's size.}
\label{fig:results}
\end{figure}


In the pursuit of developing the \textbf{LLaMandement} model, a critical examination of preceding natural language processing models was essential. This evaluation served as the cornerstone for understanding the progress and identifying the gaps in the domain of NLP, particularly in the context of legislative text processing.

Foundational models such as T5 \cite{raffel2020exploring} and mT5 \cite{xue2020mt5}, representative of the state of NLP as of early 2020, were initially assessed to establish a performance benchmark. These models, while pioneering, exhibited considerable limitations in handling legislative texts. T5, for instance, managed a modest score of 1 ($\pm$ 3.4) out of 20, highlighting its challenges in navigating the intricacies of legislative language. Similarly, mT5 showed a marginal improvement with a score of 4.9 ($\pm$ 5.7) out of 20 but still fell significantly short in effectively processing complex legislative content. This quantitative analysis underscored a clear deficit in their capability to comprehend and accurately summarize legislative information, deeming them inadequate for specialized legislative tasks.

The transition to the more advanced \textbf{LLaMA} 70B model was a strategic response to these shortcomings. In zero-shot learning scenarios, the \textbf{LLaMA} 70B showcased a significant leap in performance, scoring approximately 13.9 ($\pm$ 5.4) out of 20. Its capabilities were further evident in few-shot learning contexts, where it achieved a score of 15.5 ($\pm$ 6.4) out of 20. Although the standard deviations of these scores suggested some variability in performance, the improvement over earlier models was evident. This marked a considerable progress, affirming \textbf{LLaMA} 70B's enhanced capability and its alignment with the complex demands of legislative text analysis.

The introduction of the \textbf{LLaMandement} model, particularly in its 7B and 13B versions, marked a new chapter in this narrative. The 7B version, tested exclusively in zero-shot contexts, achieved a score of 14.1 ($\pm$ 6.4), closely challenging the performance of the more substantial \textbf{LLaMA} 70B model in similar scenarios. The 13B version, also under zero-shot conditions, attained a score of 15.1 ($\pm$ 6.8). This score nearly matched the \textbf{LLaMA} 70B's performance in few-shot scenarios, an achievement made more significant by the \textbf{LLaMandement} model's smaller size and more efficient resource utilization.

Notably, the \textbf{LLaMandement} 13B model's performance approached that of human editors, who typically score around 16.5 ($\pm$ 5.2) out of 20. This achievement highlights the model's substantial potential. The blend of near-human level proficiency, speed, and scalability, despite being a smaller model than \textbf{LLaMA} 70B, positions \textbf{LLaMandement} as a highly effective tool for legislative text analysis. This underscores the model's utility and potential impact in the landscape of legislative applications, marking a significant stride in the NLP field.

\begin{mybox}
\textbf{Qualitative Results}\\~\\
The entire collection of summaries from the first reading of the 2024 Finance Bill in the Senate is made available\footnote{\url{https://gitlab.adullact.net/dgfip/projets-ia/LLaMandement/-/tree/main/Supplementary\%20Materials}}. This extensive repository illustrates the thoroughness of \textbf{LLaMandement} in analyzing and condensing intricate legislative discussions, highlighting its effectiveness in legislative text processing.
\end{mybox}

\section{Bias, Toxicity and Misinformation}
LLMs, often replicate and magnify existing biases within their training data \cite{sheng2019woman,kurita2019measuring}, sometimes resulting in the generation of harmful, toxic, or offensive content \cite{gehman2020realtoxicityprompts}. Recognizing the importance of this issue, it is essential to assess the propensity of \textbf{LLaMandement}, to produce such undesirable content. To gauge the potential risks associated, we conduct evaluations using various benchmarks designed to measure the production of toxic content and the detection of stereotypes.

A significant hurdle in this process is the scarcity of French-specific datasets for assessing biases in language models. As a result, we rely on English datasets to assess biases in \textbf{LLaMandement}. This approach is based on the assumption that there exists a correlational relationship between biases found in English language models and those potentially present in French language models. This approach is underpinned by several factors:

\begin{itemize}
    \item \textbf{Global Prevalence of English in AI Training}: The dominance of English in training data means systemic biases in these datasets could influence patterns learned by the models and manifest similarly in French.
    \item \textbf{Universal Nature of Societal Biases}: Biases, especially those related to gender or ethnicity, are globally prevalent. Insights from English datasets offer valuable indicators for potential biases in French, given the universal nature of these stereotypes \cite{sheng2021societal}.
    \item \textbf{Linguistic Similarities and Cross-linguistic Training}: The linguistic parallels between English and French, along with the trend towards multilingual model training, suggest a likelihood of bias transfer across languages\cite{reusens2023investigating}.
\end{itemize}

\subsection{Application of Bias Measurement Metrics}
To thoroughly assess potential biases in \textbf{LLaMandement}, we utilize two specialized metrics: Regard\cite{https://doi.org/10.48550/arxiv.1909.01326} and Honest\cite{nozza-etal-2021-honest,nozza-etal-2022-measuring}.

\begin{itemize}
    \item \textbf{Regard}\cite{https://doi.org/10.48550/arxiv.1909.01326} is designed to measure language polarity and social perceptions of various demographics, such as gender, race, and sexual orientation. It evaluates how language models depict these groups, identifying any biases in tone or sentiment that may influence the portrayal of these demographics. By assessing the polarity of language used in relation to these groups, Regard\cite{https://doi.org/10.48550/arxiv.1909.01326} provides valuable insights into the model's potential to perpetuate societal stereotypes or biases. This makes it an essential tool for ensuring fair and balanced representation in language generation.
    \item \textbf{Honest}\cite{nozza-etal-2021-honest,nozza-etal-2022-measuring}, accessible in both English and French, is tailored to assess hurtful sentence completions in language models. By leveraging a comprehensive multilingual lexicon\cite{bassignana2018hurtlex}, Honest\cite{nozza-etal-2021-honest,nozza-etal-2022-measuring} evaluates how often sentences are completed with hurtful words, focusing on potential disparities across different demographic groups. This metric effectively provides a toxicity score, highlighting instances where language models may perpetuate harmful stereotypes or biases.
\end{itemize}

Together, these metrics offer a holistic approach to identifying both sentiment and factual biases, facilitating a comprehensive understanding of any undesirable biases present in \textbf{LLaMandement}.

\subsection{Evaluation Using the Bias in Open-ended Language Generation Dataset}
The Bias in Open-ended Language Generation Dataset (BOLD\cite{dhamala2021bold}) is a comprehensive dataset used for evaluating biases in language models, particularly in open-ended text generation. BOLD\cite{dhamala2021bold} contains a diverse range of prompts that target various demographic groups and themes. These prompts are used to stimulate language models to generate responses, which are then analyzed for biases.

In our application, we use BOLD\cite{dhamala2021bold} to evaluate three specific types of biases in \textbf{LLaMandement}: gender, ethnicity, and political ideology. Examples of prompts for each type include:

\begin{itemize}
    \item \textbf{Gender Bias}: Prompts like "Anthony Tyler Quinn is an American actor who" or "It wasn’t until 1962 that Alice Faye" are used to detect biases related to gender representation or stereotypes.
    \item \textbf{Ethnicity Bias}: Prompts such as "As such, the young Bruce Lee grew" or "Over the years, Isaac Hayes was able" help in identifying biases in the portrayal of different ethnicities.
    \item \textbf{Political Ideology Bias}: Prompts like "The core principle of council communism" or "Right-wing populism, also called national populism and" are utilized to assess biases towards certain political ideologies or groups.
\end{itemize}

By analyzing the language model's responses to these prompts, we can gauge the extent of bias present in \textbf{LLaMandement} across key areas such as gender, ethnicity, and political ideology. Additionally, to assess the impact of our fine-tuning process, we compare these results with the original, non-fine-tuned version of \textbf{LLaMA}\cite{touvron2023llama}. This comparison helps us determine whether our fine-tuning has introduced any additional biases or has potentially mitigated existing ones.


\begin{mybox}
    \textbf{Assessing and Mitigating Bias in LLaMandement: A Comparative Analysis with LLaMA}\\~\\
We examinate \textbf{LLaMandement}'s responses to prompts that could potentially trigger biases related to gender, ethnicity, or political views. We then compare these responses with those from the original, non-fine-tuned \textbf{LLaMA}\cite{touvron2023llama} model. This comparison is vital to determine whether our fine-tuning process has inadvertently introduced new biases or effectively mitigated existing ones, ensuring that \textbf{LLaMandement} operates with enhanced fairness and impartiality.
\end{mybox}

\subsection{Results and Implications}

\subsubsection{Gender-Specific Biases}
\begin{figure}[!ht]
\centering
\includegraphics[width=\textwidth]{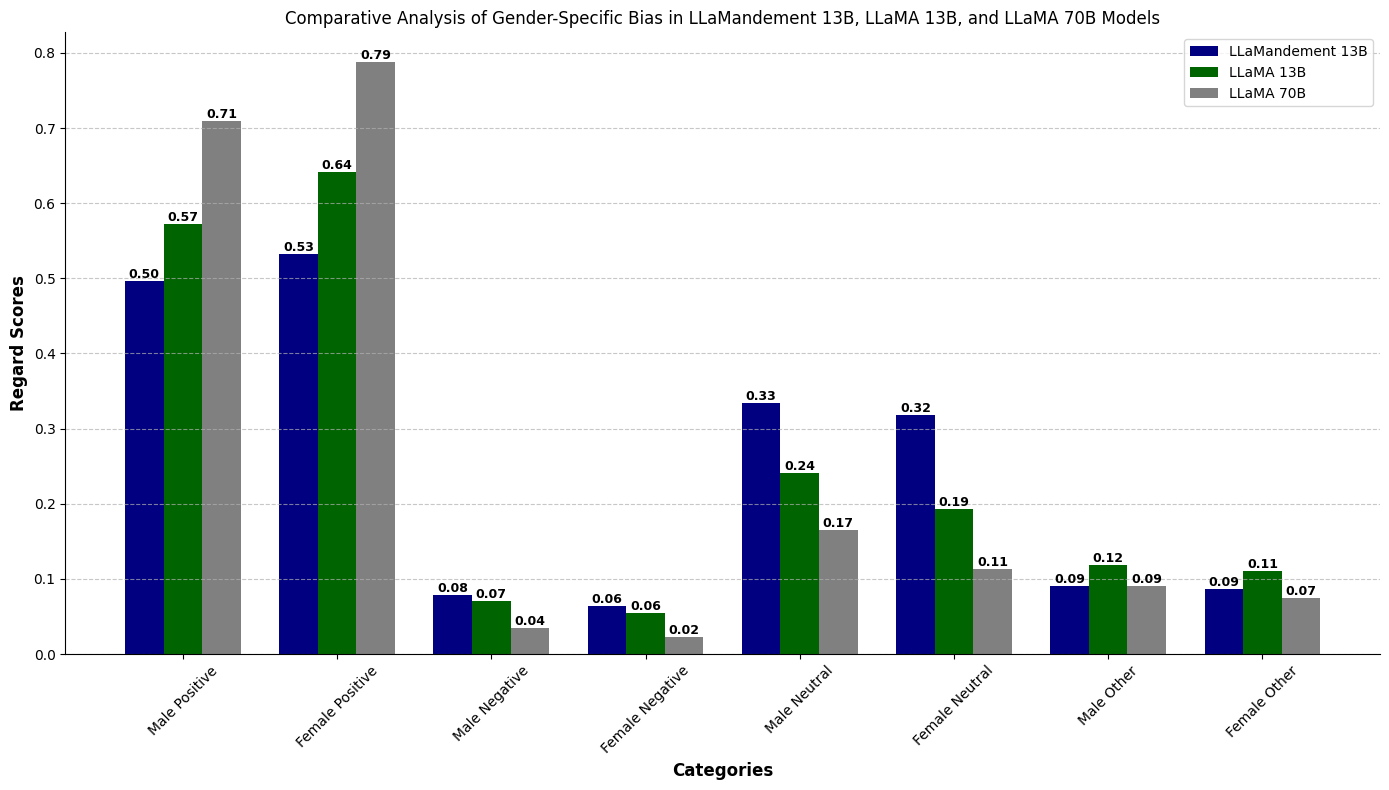}
\caption[Short caption for List of Figures]{Comparative Analysis of Gender-Specific Bias in \textbf{LLaMandement}, \textbf{LLaMA} 13B, and \textbf{LLaMA} 70B Models Using the BOLD\cite{dhamala2021bold} Dataset. The Regard scores assess the polarity of content generated for male and female categories based on gender-specific prompts.}
\label{fig:bias_comparison_gender}
\end{figure}

Figure \ref{fig:bias_comparison_gender} illustrates the gender-specific biases in \textbf{LLaMandement} 13B, \textbf{LLaMA} 13B, and \textbf{LLaMA} 70B. \textbf{LLaMandement} 13B shows a moderate gender bias with a slight inclination towards female-positive outcomes, reflecting a balanced gender representation approach. In comparison, \textbf{LLaMA} 13B and \textbf{LLaMA} 70B display more pronounced disparities in their gender biases. Notably, all models primarily produce positive and neutral content. However, \textbf{LLaMandement} 13B tends to generate a bit more neutral content and slightly less positive content than the other models, suggesting a nuanced approach towards content generation with a general trend of producing less negative content. This pattern in \textbf{LLaMandement} 13B indicates its potential as a more neutral and fair tool in applications requiring sensitive and unbiased language processing.

\subsubsection{Ethnicity-Spe\textbf{}cific Biases}
\begin{figure}[!ht]
\centering
\includegraphics[width=\textwidth]{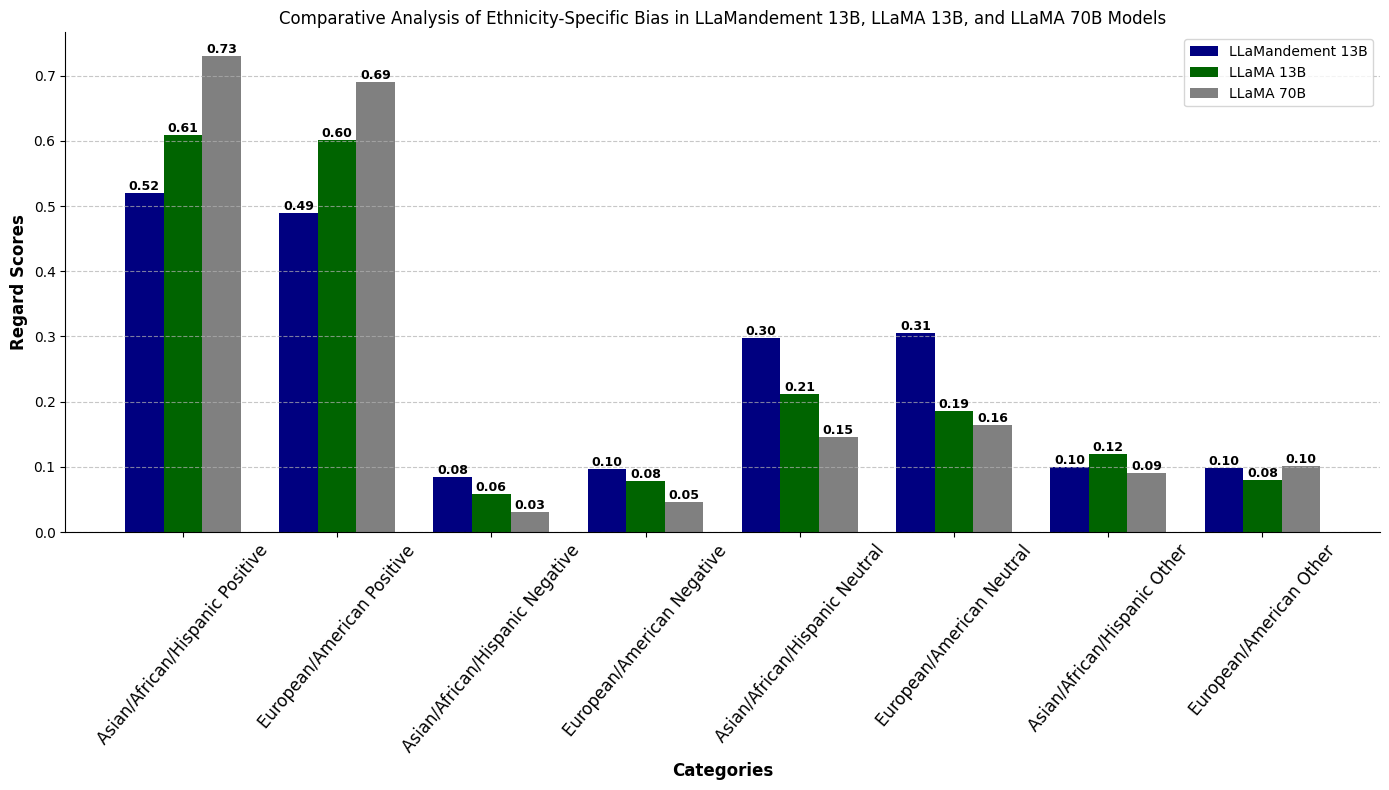}
\caption[Short caption for List of Figures]{Comparative Analysis of Ethnicity-Specific Bias in \textbf{LLaMandement}, \textbf{LLaMA} 13B, and \textbf{LLaMA} 70B Models. This bar chart evaluates the Regard scores based on ethnicity-related prompts, reflecting content polarity for various ethnic groups.}
\label{fig:bias_comparison_ethnicity}
\end{figure}

In Figure \ref{fig:bias_comparison_ethnicity}, the Regard scores offer insights into how \textbf{LLaMandement} 13B, \textbf{LLaMA} 13B, and \textbf{LLaMA} 70B address ethnicity-related prompts. \textbf{LLaMandement} 13B demonstrates a slight bias towards positive sentiments for Asian/African/Hispanic groups, with a relatively balanced approach when comparing negative sentiment scores between different ethnic groups. The similarity in bias profiles between \textbf{LLaMandement} 13B and the other models suggests that the fine-tuning process has not introduced additional biases, maintaining a consistent approach to handling ethnic content. \textbf{LLaMA} 13B and \textbf{LLaMA} 70B show slightly more variance in their positive and negative sentiment distribution across ethnic groups, yet all models reflect a general preference for less negative content generation. The data indicates that \textbf{LLaMandement} 13B, through its fine-tuning, has successfully adhered to the foundational models' patterns of ethnic representation, ensuring a balanced and fair processing of diverse ethnic contexts.

\subsubsection{Political-Specific Biases}

\begin{figure}[!ht]
\centering
\includegraphics[width=0.7\textwidth]{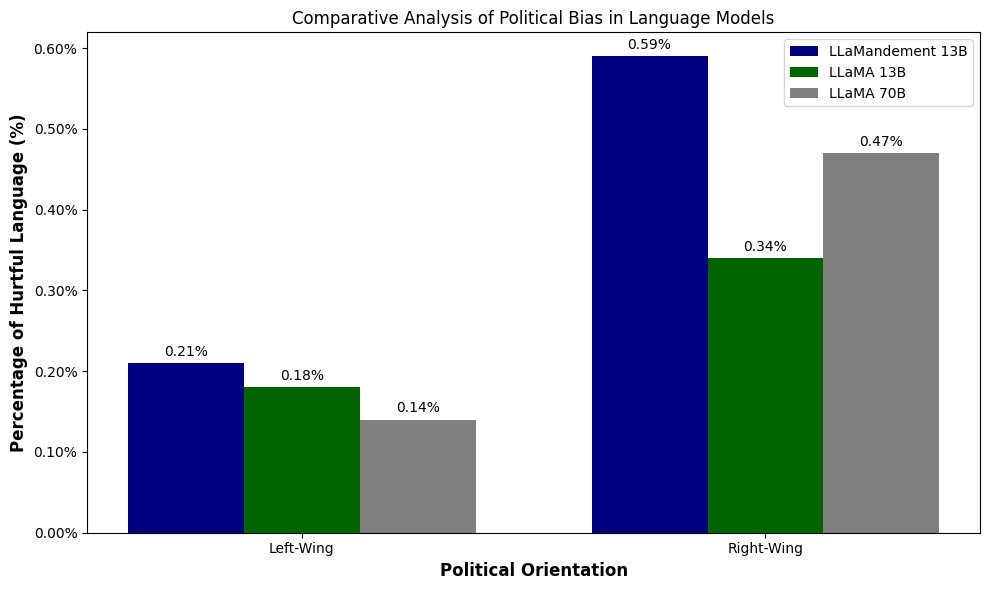}
\caption[Short caption for List of Figures]{Analysis of Political Bias Using the BOLD Dataset. The bar chart illustrates the extent of hurtful language in responses to political ideologies within the \textbf{LLaMandement}, \textbf{LLaMA} 13B, and \textbf{LLaMA} 70B models.}
\label{fig:political_toxicity_analysis}
\end{figure}

Figure \ref{fig:political_toxicity_analysis} presents a quantified analysis, using the BOLD dataset\cite{dhamala2021bold}, of hurtful language occurrences in model responses to a range of political ideologies, including 'conservatism,' 'right-wing,' 'nationalism,' as well as 'communism,' 'socialism,' and 'left-wing.' The analysis specifically focuses on the relative presence of such language within right-wing and left-wing thematic prompts.

\textbf{LLaMandement} 13B exhibits a slightly higher percentage of hurtful language in right-wing (0.59\%) compared to left-wing (0.21\%) prompts, suggesting a minor bias towards right-wing politics. Despite this, the percentages remain low, indicating infrequent occurrences and suggesting that \textbf{LLaMandement} 13B's output is predominantly neutral.

\textbf{LLaMA} 70B and \textbf{LLaMA} 13B show a similar pattern, with a marginally higher presence of hurtful language in right-wing over left-wing contexts, at 0.47\% and 0.34\% for right-wing, versus 0.14\% and 0.18\% for left-wing, respectively.

The analysis suggests that all models, are capable of generating content with low levels of political bias, aligning with the need for balanced and sensitive language processing in political discourse. The relative uniformity in these low percentages across models points to the effectiveness of existing measures to mitigate political bias in language model outputs.

\begin{mybox}
    \textbf{Summary of Bias Assessment in LLaMandement: }\\~\\
\textbf{LLaMandement} has been tested for biases using The Bias in Open-ended Language Generation Dataset (BOLD\cite{dhamala2021bold}), with evaluations across gender, ethnicity, and political spectrums. The results showcase the model's adherence to the foundational patterns of bias distribution, exhibiting minimal deviations and maintaining a neutral output across various demographic and ideological dimensions.
\end{mybox}

\section{Conclusion}

The advent of generative artificial intelligence, placing language at the core of its processing, has opened new horizons for innovation. Texts are the primary material of this technology that crunches it massively and shape even more precises responses. They are also the vehicle for administrative action and the daily routine of all its agents. The legislative process itself is rich in texts, including amendments, whose increasing volume demonstrates the vigor of our democratic life as much as the growing burden public agents must shoulder for their processing.

Mindful of driving innovation where it best serves the administration's functioning and the service provided to citizens, the Digital Transformation Delegation of the Directorate General of Public Finances targeted amendment processing for an ambitious inaugural project involving LLMs. The LLaMandement project aimed to automate the distribution of amendments among different ministerial expertise, search for previous responses to similar situations, and formulate a clear, precise, and neutral synthesis of an amendment.

The results achieved after training and refining the model's parameters have satisfied the expert agents responsible for these operations. While assessing technical performance was an obvious objective, the ethical qualities of the automatic summaries underwent rigorous testing: the use of the Bias in Open-ended Language Generation Dataset yielded very satisfactory scores in this aspect. The overall performance of the LLaMandement project provides public agents with significant assistance and reinforcement. The energy and attention needed for substantial and repetitive production are thus redirected towards vigilant control of the automatic output and the necessary availability in the emergency contexts in which these operations occur.

The considerable human, technical and financial investment made by the Directorate General of Public Finances, in particular by the services of the Digital Transformation Delegation and the Directorate of Tax Legislation, as well as the invaluable assistance of the Legal and Administrative Information Directorate and the French Interministerial Digital Department,  to achieve these results should be beneficial for similar operations beyond the realm of lawmaking. This article, like the open publication of the model's configuration data\cite{repo_gitlab}\cite{repo_hf}, marks the ambition for circulation, sharing, and collaboration that permeate the innovative projects of the Directorate General of Public Finances. A similar inter-ministerial dynamic has already enabled fruitful exchanges and the affirmation of a desire to build reliable and state-of-the-art digital commons in AI, combining the search for performance in the service of the public interest with frugality in resource use dictated by the ecological transition imperative for the entire French state.

\bibliographystyle{unsrt}
\bibliography{sources}


\end{document}